\documentclass[journal]{IEEEtran}

\usepackage{array}
\usepackage[nottoc]{tocbibind}
\usepackage{hyperref}
\hypersetup{
    colorlinks=true,
    linkcolor=blue,
    filecolor=magenta,      
    urlcolor=cyan,
}
 
\ifCLASSINFOpdf
   \usepackage[pdftex]{graphicx}
   \graphicspath{{../pdf/}{../jpeg/}}
   \DeclareGraphicsExtensions{.pdf,.jpeg,.png}
\else
   \usepackage[dvips]{graphicx}
   \graphicspath{{../eps/}}
   \DeclareGraphicsExtensions{.eps}
\fi
\usepackage{lipsum}

\begin{document}
%
\title{Deep Active Learning for Axon-Myelin \\ Segmentation on Histology Data}
%
%
%

\author{M\'elanie~Lubrano~di~Scandalea,
        Christian~S. Perone,
        Mathieu~Boudreau,
        Julien~Cohen-Adad
\thanks{The authors are with NeuroPoly Lab, Institute of Biomedical Engineering, Polytechnique Montr\'eal, Montr\'eal, QC, Canada e-mail: lubrano.mel@gmail.com}
\thanks{M. Lubrano di Scandalea is from T\'el\'ecom Paristech University, Paris, France}
\thanks{M. Boudreau is with Montr\'eal Heart Institute, Montr\'eal, QC, Canada}
\thanks{J. Cohen-Adad is with Functional Neuroimaging Unit, CRIUGM, Universit\'e de Montr\'eal, Montreal, QC, Canada}
}
\maketitle

\begin{abstract}
Semantic segmentation is a crucial task in biomedical image processing, which recent breakthroughs in deep learning have allowed to improve. However, deep learning methods in general are not yet widely used in practice since they require large amount of data for training complex models. This is particularly challenging for biomedical images, because data and ground truths are a scarce resource. Annotation efforts for biomedical images come with a real cost, since experts have to manually label images at pixel-level on samples usually containing many instances of the target anatomy (e.g. in histology samples: neurons, astrocytes, mitochondria, etc.). In this paper we provide a framework for Deep Active Learning applied to a real-world scenario. Our framework relies on the U-Net architecture and overall  uncertainty measure to suggest which sample to annotate. It takes advantage of the uncertainty measure obtained by taking Monte Carlo samples while using Dropout regularization scheme. Experiments were done on spinal cord and brain microscopic histology samples to perform a myelin segmentation task. Two realistic small datasets of 14 and 24 images were used, from different acquisition settings (Serial Block-Face Electron Microscopy and Transmitting Electron Microscopy) and showed that our method reached a maximum Dice value after adding 3 uncertainty-selected samples to the initial training set, versus 15 randomly-selected samples, thereby significantly reducing the annotation effort. We focused on a plausible scenario and showed evidence that this straightforward implementation achieves a high segmentation performance with very few labelled samples. We believe our framework may benefit any biomedical researcher willing to obtain fast and accurate image segmentation on their own dataset. The code is freely available at \url{https://github.com/neuropoly/deep-active-learning}.
\end{abstract}

\begin{IEEEkeywords}
Active Learning, Axon, Convolutional Neural Network, Deep Learning, Histology, Myelin, Segmentation. 
\end{IEEEkeywords}

%
\IEEEpeerreviewmaketitle

\section{Introduction}
\label{sec1}

In the last few years, breakthroughs have been achieved by the biomedical imaging community thanks to major advances in deep learning, especially in semantic image segmentation which is a crucial task in biomedical image processing and more specifically in neuroimaging. Being able to measure precisely the myelin density along the spinal cord can be decisive for understanding the patho-physiology of neurodegenerative diseases (e.g. multiple sclerosis, \cite{Lassmann2014-rb} or spinal cord injury \cite{Papastefanaki2015-dp}). The development of personalized therapy would benefit from precise and automatic semantic segmentation, made possible by new deep learning architectures.

Semantic segmentation boils down to assign a class label to each pixel of the image. Traditional methods take advantage of fully convolutional networks (FCN), a variant of the common convolutional neural networks (CNN), which require large number of images and ground truths to be trained. More recent approaches are making a much better use of pixel’s context information as presented in the U-Net \cite{Ronneberger2015-xy}: a contracting path is capturing the high-level context information whereas a symmetric expanding path is providing precise information about the localization. However, this method still relies on labeled data availability. Indeed, the drawback of deep learning architectures, as they involve millions of parameters, is the necessity of large amount of input data to feed the network and avoid overfitting. Yet, biomedical images are scarce and ground truths hard to obtain; since they require medical experts to manually annotate them.

\begin{figure}
    \centering
    \includegraphics[scale = 0.28]{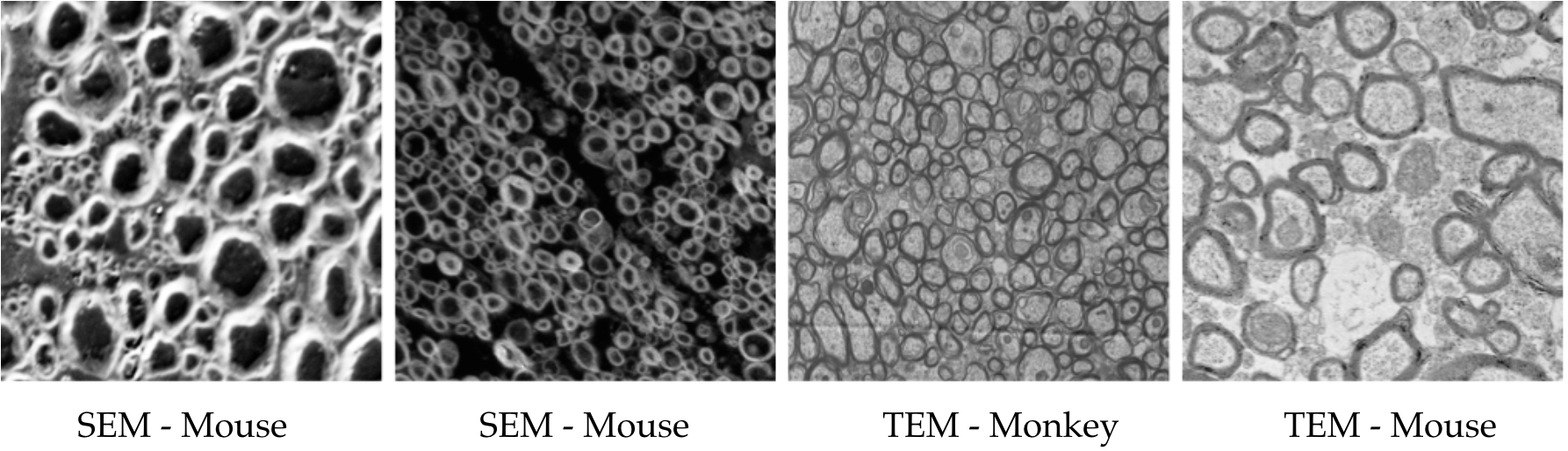}
    \caption{Data Variability for axon-myelin histologies acquired with either Scanning Electron Microscopy (SEM) or Transmission Electron Microscopy (TEM), for samples in the brain and spinal cord.}
    \label{fig:data_variability}
\end{figure}

Most of the time, real-world medical imaging datasets are considered really small for deep learning applications and center specific: for a same acquisition method, e.g. electron microscopy, images will be very different from one research lab to another or even across different subjects, as they depend on the sample preparation, its location in the body, or even the microscope settings (see Fig. \ref{fig:data_variability} for examples). Efficient software solutions such as AxonDeepSeg \cite{Zaimi2018-kj}, providing pre-trained models for specific acquisition method - Scanning Electron Microscopy (SEM) and Transmission Electron Microscopy (TEM) -  have therefore reached their limits.

However, we usually observe a certain aspect consistency within a single dataset (eg. same subject, same preparation, same microscope). Medical researchers will be willing to segment data issued from this single acquisition center. Therefore, segmentation performances would highly beneficiate from a model specifically trained for their particular dataset. Yet, a limitation arise: obtaining ground truth is an extremely tedious task and reducing the manual annotation effort by experts becomes a new priority. 

To mitigate this burden, new methods have been proposed such as transfer learning \cite{Van_Opbroek2013-qs}  or weakly and semi-supervised learning \cite{Papandreou2015-kf}. 
In the recent literature, active learning seems to be a promising and popular alternative \cite{Settles2012-ru}. The main concept relies on annotating judiciously the most informative samples. Active Learning has proven to be efficient for biomedical image segmentation. In \cite{Yang2017-td}, the authors estimate uncertainty out of FCNs using the concept of bootstrapping: \textit{a set of models is trained while restricting each of them to a subset of the training data}, the uncertainty is therefore the average pixel-wise variance among this set of model. Alternatively, Gaur et al. proposed to compute an average uncertainty score of a given region using individual pixel classification probabilities \cite{Gaur2016-sr}. 
These studies achieved similar results as the state-of-the-art methods on a given dataset, using less labeled data (down to 50-30 \% of the initial dataset sometimes). However, for both methods, the uncertainty selection criteria is not self-sufficient and authors also considered similarity measure to select new samples. The elected images to be annotated were not only the most uncertain, but the ones carrying the more novelty information among the most uncertain within the dataset. Besides, although these methods presented good results, they are often tested on large and already well curated datasets such as MNIST \cite{Lecun1998-zv} or Melanoma skin cancer dataset \cite{Isic2017-zq}, which can be considered far from a real-world scenario. Furthermore, in addition to being hard to implement, these uncertainty measures are subject to discussion: in \cite{Gal2016-sf} the author made it clear that classification probability of a network should not be considered as an estimation of the model epistemic uncertainty.

Therefore, one of the main challenges for an Active Learning framework is to evaluate which samples will be the most informative for the model. The difficulty resides then in the definition of an acquisition function that will query the appropriate sample to be annotated. Ideally we would like this acquisition function to request annotation from the human expert for the sample that will help the most the model to generalize and make accurate prediction.
In \cite{Gal2016-ij}, the authors are taking advantage of the uncertainty defined in \cite{Gal2016-sf}. They exposed how using Dropout \cite{Srivastava2014-wf} during test time, as an approximation of variational bayes estimator would allow to measure the uncertainty of the neural network.

In this work we propose an end-to-end Active Deep Learning framework for the segmentation of myelin sheath along neuronal tissue from histology data. We take advantage of Monte-Carlo Dropout (MC-Dropout) \cite{Gal2016-sf} to assess the model uncertainty and select the samples to be annotated for the next iteration.

\section{Methods}
\subsection{Active learning framework}

\begin{figure}
    \centering
    \includegraphics[scale = 0.41]{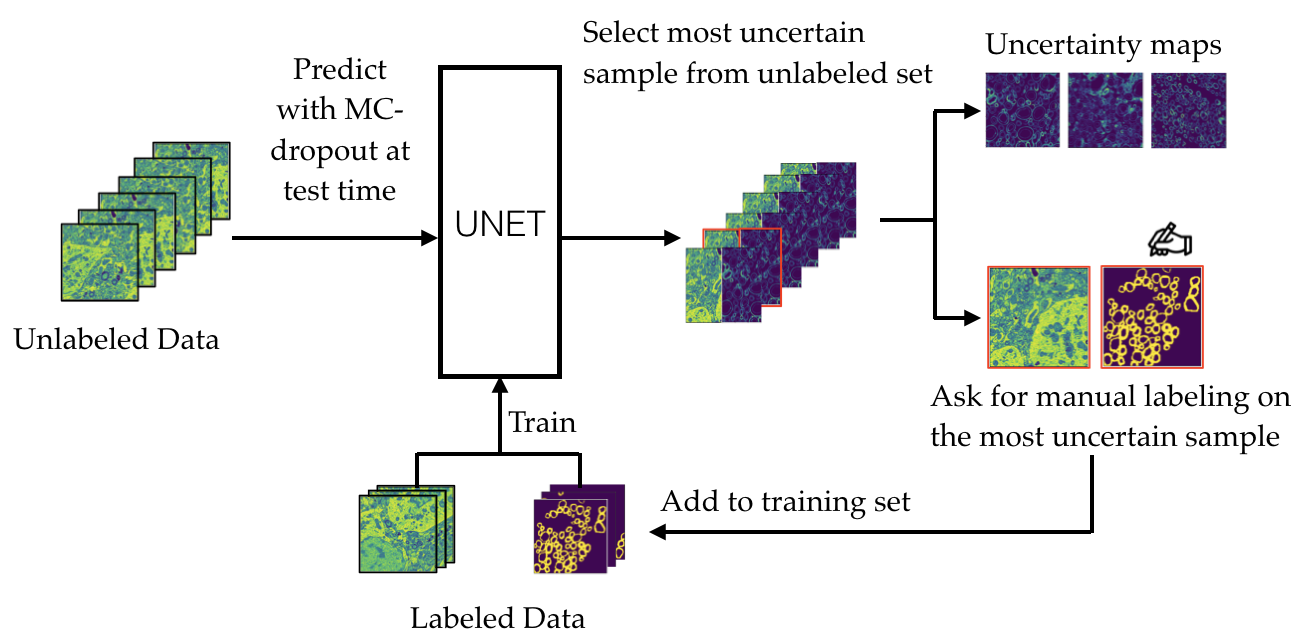}
    \caption{Active Learning pipeline for simulation: the pool of unlabeled data is fed to the U-Net multiple times, the most uncertain sample is selected to be manually annotated by a human oracle, and then added to the training set to start a new \textbf{active learning iteration}.}
    \label{fig:al_workflow}
\end{figure}

The Fig. \ref{fig:al_workflow} describe the pipeline simulating active learning iteration on the datasets.

This simulation environment was inspired by the Cost Effective Active Learning framework proposed in \cite{Blanch2017-dt}. An initial (small) labeled dataset is used to train a FCN. A pool of unlabelled images is fed into the trained U-Net and a measure of uncertainty is computed for each unlabelled sample.

They are subsequently ranked based on this uncertainty measure. The most “uncertain” samples (one or more), as defined later on, is then selected to be annotated by an oracle such as a human expert and added to the training set. The U-Net is then re-trained from scratch with this updated set of images.

This framework also allows to output additional information: uncertainty maps for each unlabelled samples. These uncertainty heat-maps are displaying the uncertainty at pixel level.

\subsection{Fully Convolutional Neural Network}
The neural network used here was based on the U-Net architecture \cite{Ronneberger2015-xy}. It is composed of one contracting path, using traditional convolution, to capture the context of the image by extracting the high-level features. One symmetric dilating path is using up-convolutions to capture the precise localization information of the image. Five levels of convolution and ReLu activations are followed by five levels of up-convolution until the output reaches the size of the input. Batch Normalization layers are also implemented before each activation.

This network design was motivated by two main constraints: first, the network should be able to perform well even on extremely small datasets, and still generalize enough when adding new data, therefore should be as less constraining as possible. Second, the training should not reach prohibitive time since the network will have to be re-trained from scratch after every active learning iteration. Therefore, a trade-off between number of epochs, size of the epoch, number of filters, size of the kernel, size of the input images and finally performances is required. 

Additionally, to perform MC-Dropout and generate stochastic MC samples while regularizing the model, layers of dropout are introduced after every MaxPooling layers.
Details of the U-Net architecture are represented in the Appendix section at the end.

\subsection{Measuring Uncertainty}

Active learning relies on the ability to select the right sample to be annotated to spare annotation time. Therefore, defining the right “acquisition function”, i.e. the criteria on which new samples will be selected, is a real challenge. 

\begin{figure}
    \centering
    \includegraphics[scale = 0.28]{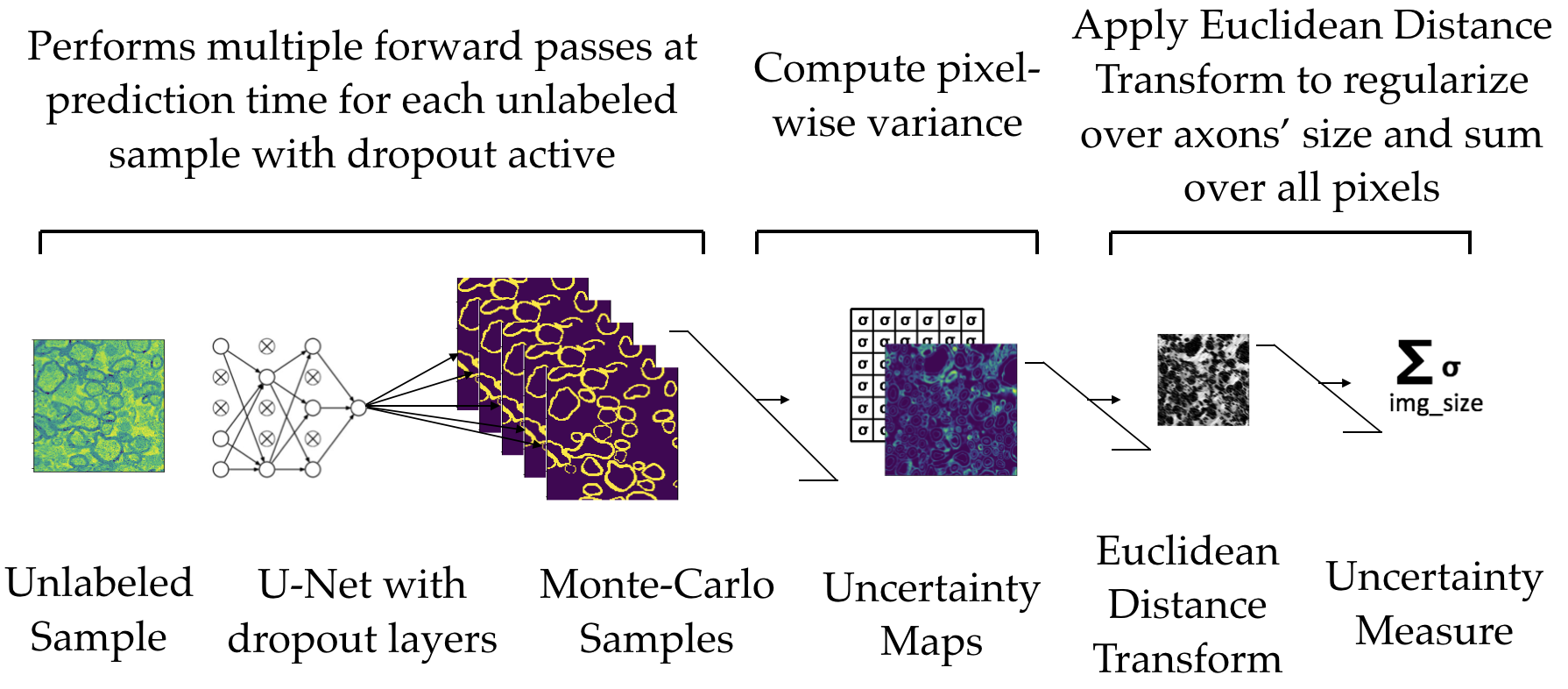}
    \caption{Uncertainty measure: uncertainty is measured for each unlabeled sample. Multiple predictions (MC-Samples) are generated from the previously-trained U-Net. They all differ thanks to the various dropout configurations. Standard deviation across all MC-Samples is then computed for each pixel. The heat-map of pixel-wise standard deviation (uncertainty map) is then weighted by the Euclidean Distance map (computed on averaged MC-samples predictions) in order to normalize by the axon size. Uncertainty measure is finally obtained by averaging the final pixels’ values.}
    \label{fig:uncertainty_measure}
\end{figure}

So far, uncertainty in neural networks has been characterized in many different ways. However, we owe a popular definition to Yarin Gal \cite{Gal2016-sf}: the confidence of a model on a sample prediction can be obtained with a Bayesian equivalent of a FCN. It has been shown that Bayesian FCN model can approximate variational inference by taking advantage of stochastic regularization techniques such as dropout. Indeed, by keeping active the dropout at prediction time and by performing multiple forward passes we can sample from the approximate posterior. The multiple predictions will be slightly different as different neurons will be activated or deactivated thanks to the dropout stochasticity. This method is referred as Monte-Carlo Dropout (MC-Dropout). In our experiments, we realized 50 forward passes at prediction time for each unlabelled image to obtain the 50 MC-samples. The uncertainty is then defined as the posterior probabilities' standard deviation of the 50 predictions. The overall uncertainty is then computed by summing uncertainty maps' pixels values. The number of MC-samples has been chosen to balance the trade-off between having a meaningful standard deviation (generating enough samples) while not increase too much the prediction time. See Fig. \ref{fig:uncertainty_measure} for more details. 

Summing the standard deviation over the image can lead to a biased measure of uncertainty. The model seems more uncertain near class borders, thus, an image containing more axons could have a higher overall variance but still be perfectly segmented. To overcome this issue, we propose to multiply the uncertainty map image with the Euclidean Distance transformation of the prediction (0.5 threshold applied to prediction probabilities). This transformation will make the border pixels less important than deeper pixels in the myelin sheath and therefore, tends to attenuate border prediction errors. 

\subsection{Datasets and Ground Truth labelling}
The first dataset consisted of 14 slices-images of an adult mouse striatum volume via \textit{serial block-face electron microscopy} (SBEM) \cite{Mikula2015-zj}. It highlights microstructures such as axons and myelin sheath within the white-matter of the brain. The slices were part of total volume represented by 1000 slices, and were selected uniformly inside the striatum volume. Each slice (image) was cut into patches of 512x512 pixels to be fed into the network afterwards. The pixel size is 20x20 nm. 

\begin{figure}
    \centering
    \includegraphics[scale = 0.26]{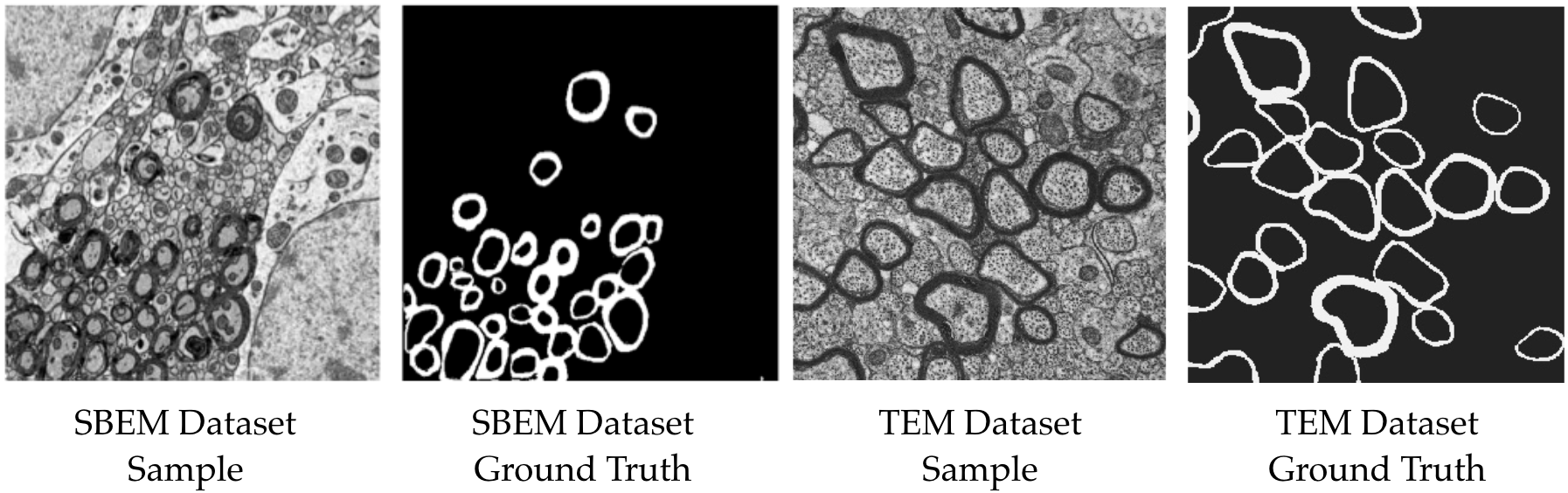}
    \caption{Samples and Ground Truth of SBEM and TEM datasets}
    \label{fig:data_samples}
\end{figure}

The second dataset contained 96 sub-patches of 512x512 pixels of multiple mice spinal-cord acquired using \textit{Transmission Electron Microscopy} (TEM) technique and sampled from 24 images with a resolution of 2.4 nm per pixel. This dataset is freely available from the \textit{White Matter Microscopy Database} \cite{Cohen-Adad_Does_DUVAL_2019}. The ground truth labelling was manually created (using image editing software such as GIMP). All ground truth labels were cross-checked by at least two researchers. Some samples and associated ground truth labels are represented in Fig. \ref{fig:data_samples}. 

\subsection{Training}
\subsubsection{Training Procedure}
Binary class segmentation is performed on both of these datasets: the myelin sheath is segmented from any background pixels. One active learning iteration corresponds to a full training on the training set. Once the most uncertain sample is selected, it is added to the training set and the network is re-train from scratch on the updated training set. The validation set remains the same during the training as well as the test set, on which the segmentation performances are computed. Fig. \ref{fig:al_workflow} illustrates this workflow.

\subsubsection{Data Augmentation}
Since the network is trained on a small training set that will increase over time, an appropriate data augmentation strategy is essential to limit overfitting and improve generalization. Therefore, augmentations such as random shifting, rotation, flipping, zooming with small ranges are included. Details are in Table \ref{table:da}. 

\begin{table}[h!]
\caption{\label{table:da} Data Augmentation strategy}
\centering
\begin{tabular}{|p{2.25cm}|p{5cm}|}
\hline
\centering \textbf{Data Augmentation parameter} & \textbf{Description}\\
\hline
Shifting & Random horizontal and vertical shifting between 0 and 10\% of the image’s size. \\ \hline
Rotation & Random rotation between -10 and 10 degrees.                                     \\ \hline
Flipping & Random horizontal and vertical flipping.                                        \\ \hline
Zooming & Random zooming with a random factor between 1/1.2 and 1.2 .                     \\
\hline
\end{tabular}
\end{table}

\subsubsection{Hyper-parameters}
Considering the small size of the training set, the network’s hyper-parameters should be as conservative as possible. The Dice loss and weighted binary cross-entropy were successively used to train the U-Net. This hyper-parameters are not re-optimized after every active learning iteration, i.e. as the training size is growing. 

Several values of hyper-parameters for both SBEM and TEM datasets training have been tested, the settings providing the most satisfying results are summarized in Table \ref{table:hp}

\begin{table}[h!]
\caption{\label{table:hp} Training Hyper-parameters}
\centering
\begin{tabular}{|p{3cm} | p{2.2cm} | p{2.2cm}|} 
\hline
\centering \textbf{Hyper-parameters} & \textbf{SBEM Dataset} & \textbf{TEM Dataset }\\ 
\hline\hline
\centering input size & 512x512  & 512x512  \\ 
\hline
\centering batch size & 5 & 5\\
\hline
\centering epochs & 3500 & 800 \\
\hline
\centering steps per epoch & 2 & 2 \\
  \hline
 \centering dropout & 0.2 & 0.2 \\
  \hline
 \centering learning rate & 1e-2 & 1e-2  \\
  \hline
 \centering decay rate & learning rate / number of epochs & learning rate / number of epochs \\
  \hline
\centering  optimizer & ADAM & ADAM  \\
  \hline
\centering  loss function & Dice loss & Dice loss  \\
  \hline
\centering  activation & sigmo\"{i}d & sigmo\"{i}d \\[1ex]
 \hline
 \centering  activation threshold & 0.5 & 0.5 \\[1ex]
 \hline
\end{tabular}
\label{table:hp}
\end{table}

\subsubsection{Loss functions}
We tested two functions: the Weighted Binary Cross-entropy (Weighted BCE) loss and the Dice loss. The Weighted BCE is a modified version of the classic Binary Cross-entropy (BCE) where weights are applied to correct class imbalance. 

The BCE loss increases as the predicted probability diverges from the actual label. It is defined as follow:

\begin{equation}
  \mathcal{L}(y_{true}, y_{pred}) =  - y_{true} \log(p_{pred})+(1 - y_{true})\log(1 - p_{pred}))
\end{equation}
where $y_{true}$ is the binary indicator ($0$ or $1$) of the class label and $p_{pred}$ the predicted probability. 

The \textbf{Weighted BCE} is therefore computed by multiplying the binary crossentropy with the weight vector:
\begin{equation}
  \theta =  0.30*y_{true} + 0.70*(1 - y_{true})
\end{equation}
where $\theta$ is the Weighted BCE.
The associated weights are computed based on a pre-analysis of the data classes proportion.

The \textbf{Dice} loss, which performs better at class imbalanced problems, is computed as follow: 
\begin{equation}
     \mathcal{L}(y_{true}, y_{pred}) = 1 - \frac{2(y_{true} \cap y_{pred})}{|y_{true}| + |y_{pred}|}
\end{equation}

\subsubsection{Evaluation Metrics}
To evaluate the performances of the model, the Dice coefficient is computed between the prediction and the ground truth mask on the test set. The Dice coefficient is a popular metric for assessing the quality of a segmentation. Considering two binary images $y_{pred}$ and $y_{true}$, we define the Dice coefficient as follows: 
\begin{equation}
Dice = \frac{2(y_{pred}\cap y_{true})}{|y_{pred}| + |y_{true}|}
\end{equation}
Where $y_{pred} \cap y_{true}$ represents the intersection of two images (myelin pixels in both images), $|y_{pred}|$ the total number of myelin pixels in $y_{pred}$ and $|y_{true}|$ the total number of myelin pixels in $y_{true}$.

\section{Results}

\subsection{Active Learning and Segmentation}

A summary of the experiments settings for both datasets is available in Table \ref{table:es}

\begin{table}[h!]
\caption{\label{table:es} Experiments settings}
\centering
\begin{tabular}{| p{4.5cm} | p{1.5cm} | p{1.5cm}|} 
\hline
\centering \textbf{Settings} & \textbf{SBEM Dataset} & \textbf{TEM Dataset } \\ 
\hline\hline
Total number of images & 14  & 24  \\ 
\hline
Total number of patches & 51 & 96\\
\hline
Number of patches in the initial training set & 5 & 2 \\
\hline
Number of patches in the validation set (fixed across all experiments) & 10 & 2 \\
  \hline
Number of patches in the test set (fixed across all experiments) & 10 & 20 \\
  \hline
Remaining patches for the unlabelled pool & 26 & 72\\
\hline
Number of patches added after each active learning iteration & 1 & 1 \\
  \hline
Number of active learning iterations for each experiment & 15 & 15 \\
  \hline
Number of experiments (for averaged results) & 10 & 3  \\
  \hline
Initialization of the network after each active learning iteration & Random & Random  \\
  \hline
 Number of MC samples & 50 & 50 \\[1ex]
 \hline
\end{tabular}
\label{table:es}
\end{table}

\subsubsection{SBEM Dataset}
\begin{figure}
    \centering
    \includegraphics[scale = 0.35]{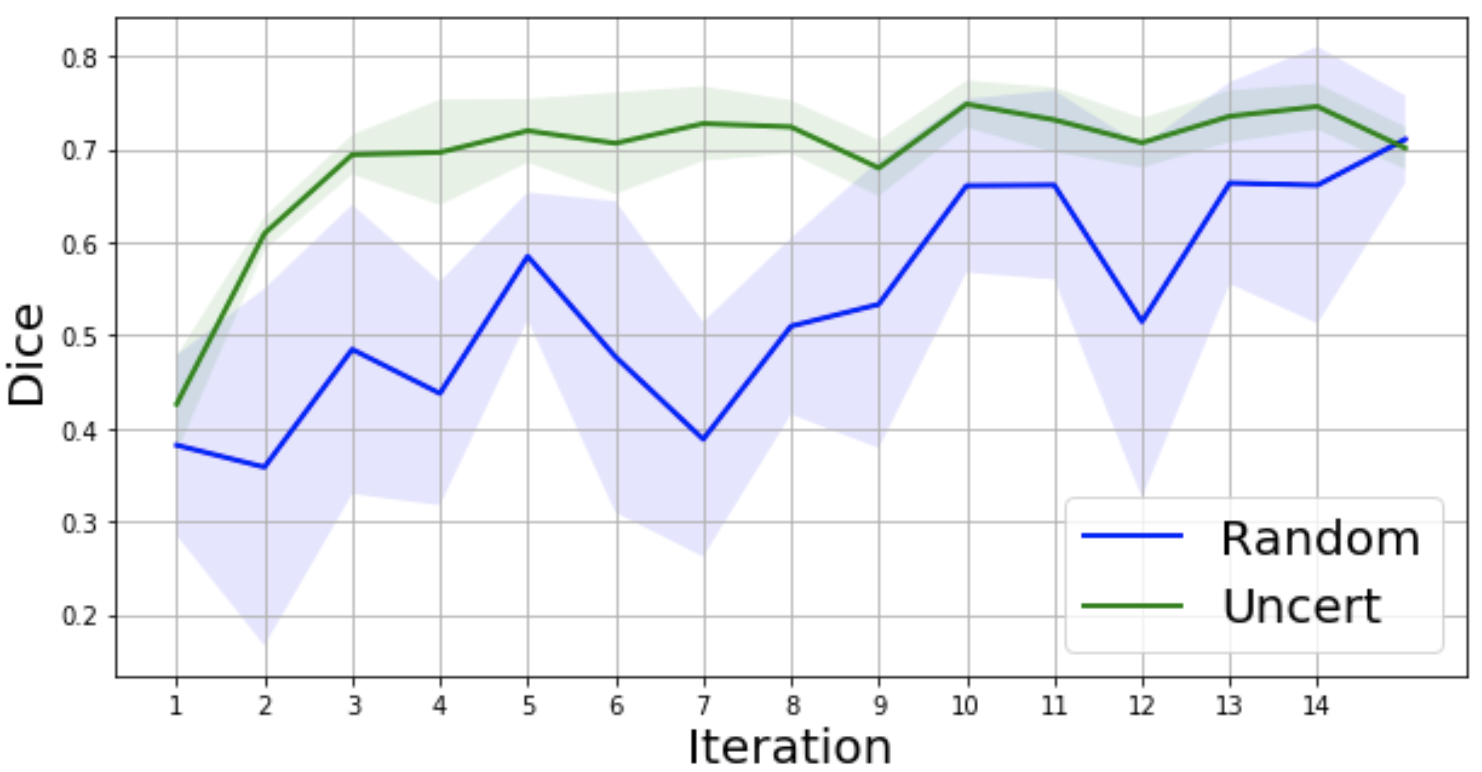}
    \caption{Active Learning simulation: Dice coefficient on SBEM test set over 15 active learning iterations (average + standard deviation over 10 experiments). One patch is added to the training set after each iteration, and the network is re-trained from scratch.\textit{"Random"} represents the baseline: each sample is selected randomly, while \textit{"Uncert"} represents the Dice obtained by adding specifically selected samples. Both experiments where trained using the Dice loss }
    \label{fig:results_mikula}
\end{figure}

The initial training set contained 5 patches. The validation set and test set (10 patches each) remained the same during all experimentations. The pool of unlabelled samples contained the 26 leftover patches. The size of each patch was 512x512 pixels. The uncertainty was computed using 50 MC-samples. A total of 15 active learning iterations was performed. Indeed, the process of selecting the most uncertain sample, adding it to the training set and retrain the U-Net from scratch was performed 15 times (therefore, 15 new samples from the pool of unlabelled data were progressively added to the training set). We ran the experiment 10 times in order to average the results. The Fig. \ref{fig:results_mikula} illustrates the mean and standard deviation of the segmentation performances evaluated on the test set across the 10 experiments. Each training phase was about 15 minutes on 2 NVIDiA Tesla P100 GPUs, and therefore, the total duration for the 10 experiments was: 10 experiments*15 iterations*15 minutes = 37 hours. As baseline, we compared our method with random selection instead of uncertainty-based selection to increment the training set. 

These results suggest a clear improvement of the proposed uncertainty method compared to the random baseline. In the early iterations, the benefits carried by uncertainty-based active learning are even more noticeable: after adding only 3 uncertainty-selected samples, the segmentation performances reaches a level that will only be obtained after adding 15 randomly-selected samples to the training set. Additionally, the gap between the two Dice curves is progressively decreasing as more samples are added. This is also consistent with the network sensibility to training set size (the more data the better). 

We observed consistency among the selected samples for each active learning iteration across the five experiments. Indeed, we would expect the uncertainty measure to be sufficiently stable to select samples in the same order when running the active learning simulation multiple times, with the same settings. By analyzing the samples selected across all the experiments, we noticed that the number of unique samples selected was much smaller when it is using the uncertainty query function than random selection. 

\subsubsection{TEM Dataset}
\begin{figure}[!h]
    \centering
    \includegraphics[scale = 0.45]{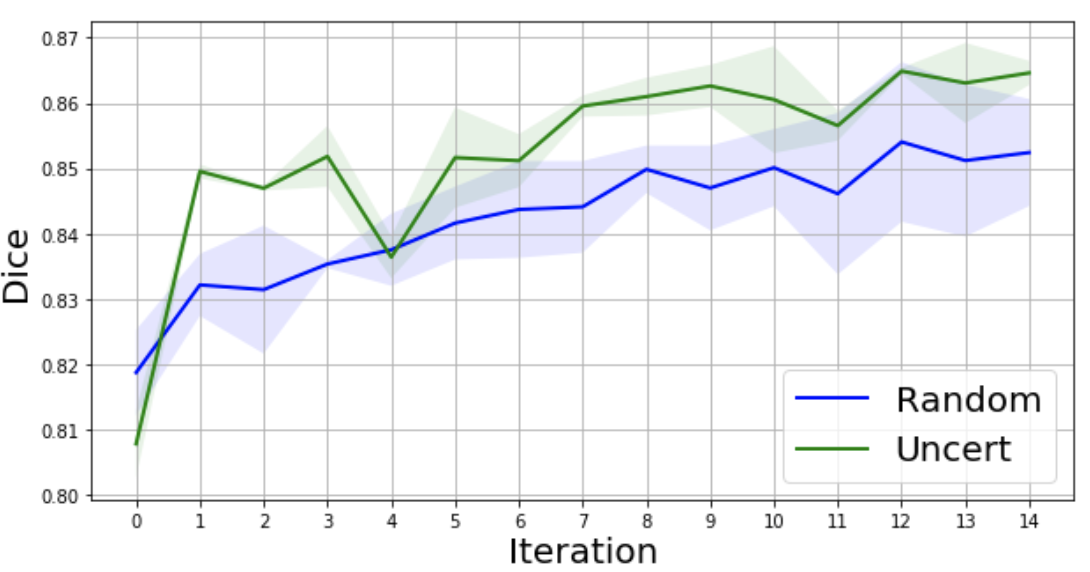}
    \caption{Active Learning simulation: Dice coefficient on TEM test set over 15 active learning iterations (average + standard deviation over 3 experiments). Again, one patch is added to the training set after each iteration, and the network is re-trained from scratch. \textit{"Random"} represents the baseline: each sample is selected randomly, while \textit{"Uncert"} represents the Dice obtained by adding specifically selected samples. Both experiments where trained using the Dice loss }
    \label{fig:results_tem}
\end{figure}

An even more extreme scenario was tested on the TEM dataset: the initial training set consists only of 2 patches, the validation set of 2 patches and the test set of 20 patches. Seventy-two patches remained for the unlabeled pool of data. Due to time and computing resources constraints we performed 15 active learning iterations per experiment and run 3 experiments to average the results. See Fig. \ref{fig:results_tem} for the results.

Results also show a clear improvement of the uncertainty-based method, with an averaged Dice value about 1.2\% higher than the random baseline at each iteration.

\subsection{Uncertainty Maps and Loss Functions}
\begin{figure*}[h!]
    \centering
    \includegraphics[scale = 0.5]{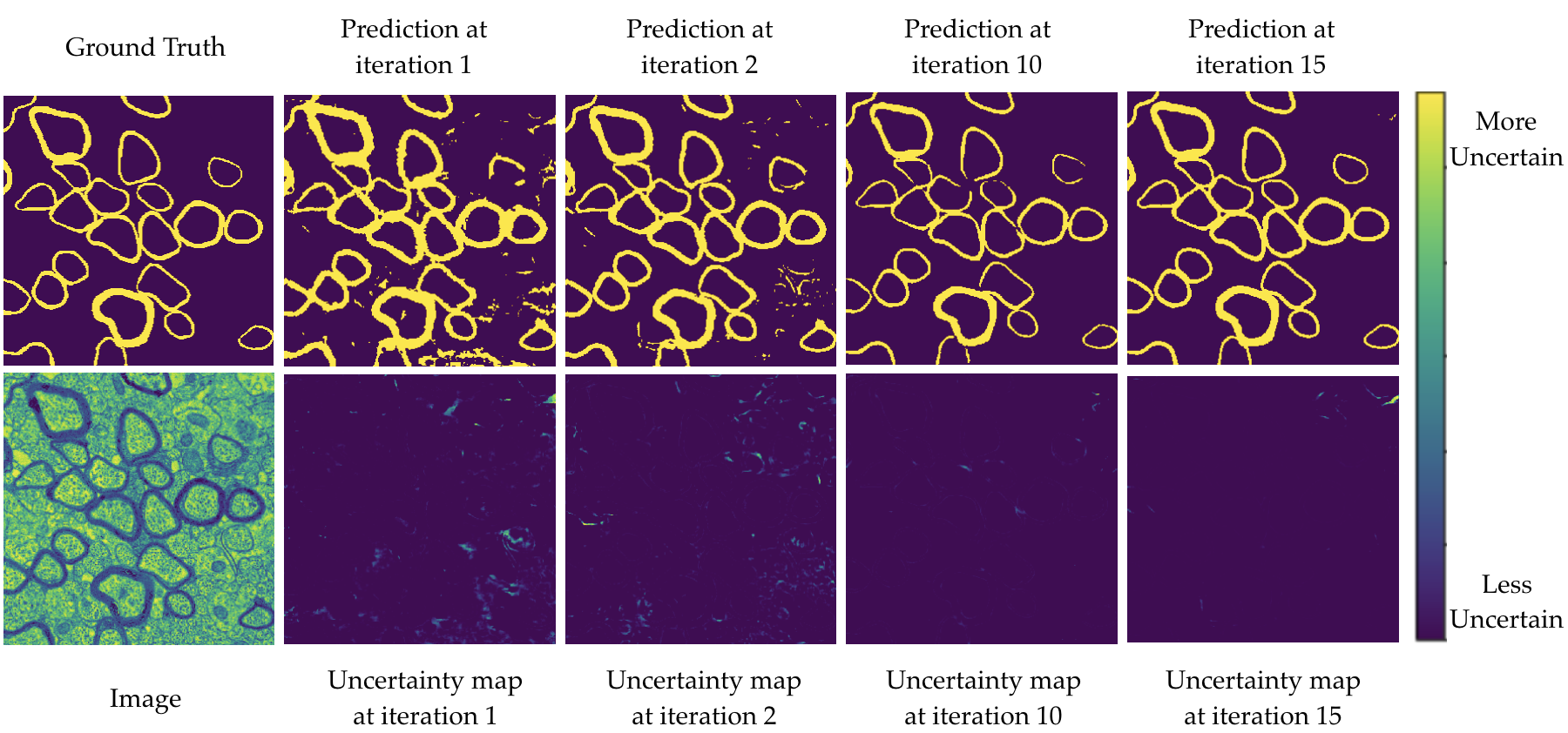}
    \caption{Uncertainty maps evolution over active learning iterations for the Dice loss function.}
    \label{fig:maps_dice}
\end{figure*}

\begin{figure*}[h!]
    \centering
    \includegraphics[scale = 0.5]{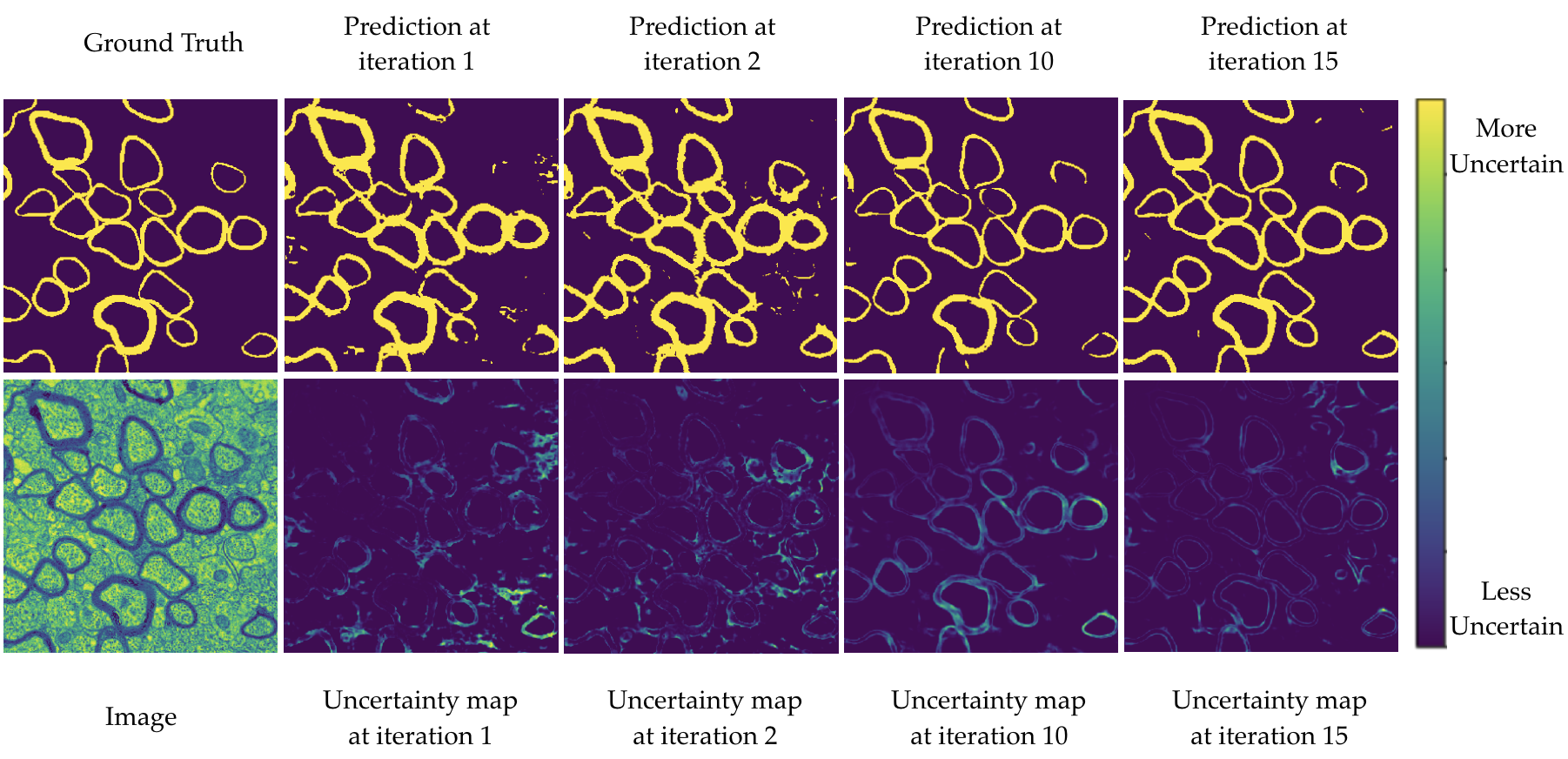}
    \caption{Uncertainty maps evolution over active learning iterations for the Weighted binary cross-entropy loss function.}
    \label{fig:maps_crossentropy}
\end{figure*}

Our implementation also outputs uncertainty maps, which could help for further understandings of intrinsic uncertainty in deep learning networks. The value of each pixel corresponds to the standard deviation computed on MC samples. The appearance of the uncertainty map depends on the loss function chosen to train the network. Indeed, the Weighted BCE seemed to lead to high uncertainty mainly on class borders and backgrounds while the Dice loss seemed to highlight contrasted areas of the image. Fig. \ref{fig:maps_dice} and Fig. \ref{fig:maps_crossentropy} illustrate the different aspects of the uncertainty maps depending on which loss function is used to train the network. Those maps are giving relevant information about areas and features on which the model tends to fail. For instance, in the case of Weighted BCE, the network seems less confident about the myelin sheath thickness and high variance is concentrated on borders pixels. On the other hand, when the network is trained with a Dice loss, the model seems to be more uncertain where it can distinguish round shapes in the background (due to other micro-structures present in the striatum for example).

\section{Discussion and Perspectives}
In this work, we implemented a framework for performing active deep learning and applied it for segmenting histology data. This approach has proven to be efficient for reducing manual labeling time when training new models on a variety of datasets. As shown here, only a reduced number of image patches can be sufficient to train efficiently a model. Indeed, the uncertainty-based selection criteria seems to select the most informative samples for the model to learn from.

\subsection{Annotation time gain}
The final goal of this study is to limit the annotation efforts required by human experts when it comes to analyzing biomedical images, and more specifically, when performing semantic segmentation. Indeed, biomedical images can be extremely complex, containing sometimes hundreds or thousands objects to annotate. For instance, if we are considering that annotating 1 axon takes about 3 minutes, 1 patch contains about 25 axons, it should take 75 minutes to annotate only 1 patch. In our case, it would have taken about 120 hours to annotate the TEM dataset, without even considering the double checking by several researchers. Therefore, every attempts to reduce human effort while preserving segmentation quality might be helpful. In this case, it has been shown that less \textbf{human} efforts was required to obtain a good segmentation. Even if the repeated network's trainings can lead to extended duration, it might still be beneficial since it is only a \textbf{computer} running. However, future works could evaluate the annotation time gain on other biomedical data such as IRM, scanners, or other microscopic images (e.g. cells). 

\subsection{Training Procedure}
As samples are added to the training set after each active learning iteration, the validation set is not filled with new samples (it is fixed for comparison purposes). This could lead to bias when the training set and validation set are too unbalanced (in the second experiment, the validation set is always containing 2 patches while the training set reaches a size of 15 patches). It might explain why the Dice curve is sometimes dropping. A solution would be to implement cross-validation so all the patches will be seen in the training set, and the validation set would also increase with time.

Another source of bias could come from the distribution of patches from one source-image to another across the training, validation and test sets. To overcome this issue we decided to fix the patches contained in the initial training set as well as in the validation and test set for all the experiments in order to specifically observe the variations caused by the new patches added. 

Furthermore, we retrained the networks from scratch after adding new data. A more efficient solution would make use of pre-trained models to initialize a network and compute the first round of uncertainty, then performing fine-tuning with the selected samples.

\subsection{Evaluation Metrics}
Another possible source of bias is related to the sole use of the Dice Coefficient to evaluate the predicted segmentation. Indeed, the Dice coefficient measures the extent of spatial overlap between two binary images but does not assess the "purity" or the "completeness" of the prediction. In future works, additional segmentation metrics could be integrated, such as specificity, sensitivity, or even accuracy.

\subsection{Uncertainty Measure}

To select the most uncertain samples to be annotated and added to the training set, we used the popular and easy-to-implement MC-Dropout uncertainty measure, highlighted in \cite{Gal2016-sf}. However, criticism of dropout uncertainty exist: Ian Osband characterizes it as \textit{approximations to the risk given a fixed model} and proposes a novel uncertainty evaluation based upon \textit{smoothed bootstrap sampling} \cite{Osband2016-mo}. Despite this, MC-Dropout demonstrated to be efficient and straightforward for our application, exploring new acquisition functions and uncertainty/ risk evaluations could improve the results as well as help understanding neural networks learning.

\subsection{Uncertainty Maps and Loss Functions}
We observed that the Dice loss and Weighted Binary Cross-entropy led to two different kind of uncertainty evaluation, as seen on the uncertainty heatmaps. The nature of neural networks’ uncertainty is different depending on the loss function they are trained with. Future work could evaluate what would be the best metric to use depending on the type of application. For instance, do we want our model to be more accurate along border pixels, even though it is detecting more false-positive in background areas?

\subsection{Software}
The ultimate objective was to provide a user friendly interface to dynamically annotate the selected samples as active learning simulation is running and integrate it in our AxonDeepSeg framework, an automatic axon-myelin segmentation tool for microscopy data using convolutional neural network \cite{Zaimi2018-kj}. This is a direction to pursue for future works. 

\section{Conclusion}
This study demonstrated an active learning framework for biomedical image segmentation. It provides an evaluation of Monte-Carlo dropout uncertainty measure, customized for real-world scenario. So far, the state-of-the-art segmentation algorithm required numerous samples to be correctly trained. This work shows that by specifically selecting a couple of highly informative samples, segmentation performances can be significantly improved.

Additionally, our study revealed interesting results regarding the visualization of uncertainty: by displaying heatmaps of this uncertainty, the user could apprehend where their model would tend to fail the most. This could be helpful to leverage the obtained predictions, especially for biomedical applications, for which quantifying and controlling the uncertainty is essential.

The code and data used in this project are freely accessible at \url{https://github.com/neuropoly/deep-active-learning}.

\section*{Appendix: U-Net architecture}
\begin{figure*}[h]
    \centering
    \includegraphics[scale = 0.45]{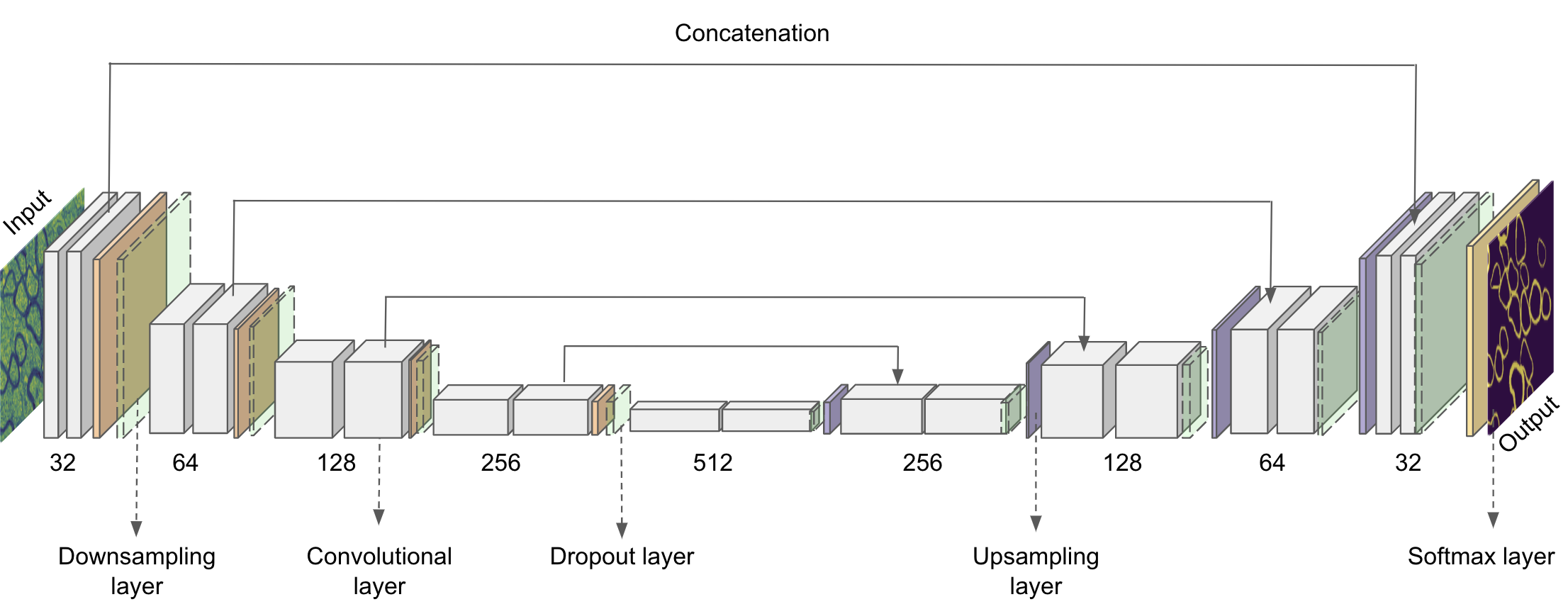}
    \caption{U-Net architecture used for these experiments.}
    \label{fig:UNET_architecture}
\end{figure*}

See Fig. \ref{fig:UNET_architecture}.

\section*{Acknowledgments}
The authors would like to thank Oumayma Bounou for fruitful discussions and for manually labelling the ground truths of SBEM data samples, Alexandru Foias for helping the authors to use GPU computation units, Dr Shawn Mikula for providing the SBEM histology Mice samples, Dr. Els Fieremans for sharing TEM data of mice. This study was funded by the Canada Research Chair in Quantitative Magnetic Resonance Imaging [950-230815], the Canadian Institute of Health Research [CIHR FDN-143263], the Canada Foundation for Innovation [32454, 34824], the Fonds de Recherche du Qu\'ébec -Santé́ [28826], the Fonds de Recherche du Québec - Nature et Technologies [2015-PR-182754], the Natural Sciences and Engineering Research Council of Canada [435897-2013], the Canada First Research Excellence Fund (IVADO and TransMedTech) and the Quebec BioImaging Network [5886].

\bibliographystyle{unsrt}
\bibliography{biblio}

\end{document}